\documentclass{article}

\usepackage[preprint]{aria_2024}
\usepackage{palatino}
\usepackage{graphicx}
\usepackage{colortbl}
\usepackage{pgf}
\usepackage{subfigure}

\usepackage{xcolor} 

\definecolor{lightorange}{rgb}{1.0, 0.9, 1.0}  
\definecolor{lightgreen}{rgb}{0., 0.66, 0.}  

\newcommand{\rgb}[1]{%
  \pgfmathsetmacro{\percent}{%
    (max(min(#1, 80), 30) - 30) / (80 - 30) * 100
  }%
  \edef\temp{\noexpand\cellcolor{lightgreen!\percent!lightorange}}\temp}
  
\usepackage{amsmath}
\usepackage{bbm}
\usepackage{arydshln}
\usepackage{multirow}
\usepackage{enumitem}
\usepackage{wrapfig}
\usepackage{makecell}
\usepackage{pgfmath}
\usepackage{pgffor}
\usepackage{float} 

\definecolor{deepgreen}{RGB}{0,150,0} 
\definecolor{deepred}{RGB}{150,0,0} 

\usepackage{pifont}

\newcommand{\name}{\textsc{Aria}}
\newcommand{\smallname}{\textsc{Aria}}

\newcommand{\eg}{\textit{e.g.}}

\usepackage[utf8]{inputenc} 
\usepackage[T1]{fontenc}    
\usepackage[colorlinks=true, linkcolor=black, urlcolor=blue, citecolor=blue]{hyperref}
\usepackage{url}            
\usepackage{booktabs}       
\usepackage{amsfonts}       
\usepackage{nicefrac}       
\usepackage{microtype}      
\usepackage{xcolor}         
\usepackage[misc,geometry]{ifsym}

\usepackage{graphicx}
\usepackage{array}
\usepackage{amsmath}
\usepackage{listings}

\renewcommand{\ttfamily}{\fontfamily{pcr}\selectfont}

\lstdefinestyle{mystyle}{
    language=Python,
    basicstyle=\ttfamily\tiny,
    keywordstyle=\color{blue},
    commentstyle=\color{deepgreen},
    stringstyle=\color{deepred},
    breakatwhitespace=false,         
    breaklines=true,                 
    captionpos=b,                    
    keepspaces=true,                 
    showspaces=false,                
    showstringspaces=false,
    showtabs=false,                  
    tabsize=2,
    morekeywords={assert,True,False}
}

\title{\name~: An Open Multimodal Native Mixture-of-Experts Model}

\author{Dongxu Li, Yudong Liu, Haoning Wu, Yue Wang, Zhiqi Shen, Bowen Qu, Xinyao Niu, \and Fan Zhou, Chengen Huang, Yanpeng Li, Chongyan Zhu, Xiaoyi Ren, Chao Li, Yifan Ye, \and Peng Liu, Lihuan Zhang, Hanshu Yan, Guoyin Wang, Bei Chen, Junnan Li\textsuperscript{\Letter} \AND Rhymes AI}

\begin{document}

\maketitle

\let\thefootnote\relax\footnotetext{\Letter~corresponding author: junnanli@rhymes.ai}

\begin{abstract}
Information comes in diverse modalities.
Multimodal native AI models are essential to integrate real-world information and deliver comprehensive understanding.
While proprietary multimodal native models exist,
their lack of openness imposes obstacles for adoptions, let alone adaptations.
To fill this gap,
we introduce \name,
an open multimodal native model with best-in-class performance across a wide range of multimodal, language, and coding tasks.
\name~is a mixture-of-expert model with 3.9B and 3.5B activated parameters per visual token and text token, respectively.
It outperforms Pixtral-12B and Llama3.2-11B,
and is competitive against the best proprietary models on various multimodal tasks.
We pre-train \name~from scratch following a 4-stage pipeline,
which progressively equips the model with strong capabilities in language understanding, multimodal understanding, long context window, and instruction following.
We open-source the model weights along with a codebase that facilitates easy adoptions and adaptations of \name~in real-world applications.

\vspace{1ex}
Code: \url{https://github.com/rhymes-ai/Aria}

Website: \url{https://rhymes.ai/}
\end{abstract}

\section{Introduction}
\label{sec:intro}
In this report, we present \name, the first open mixture-of-experts (MoE) model that is \textbf{multimodal native}. 
The term multimodal native has been used in prior literature to refer to different model capabilities, without a clear consensus.
Here, we provide a quantifiable definition:
\textit{A multimodal native model refers to a single model with strong understanding capabilities across multiple input modalities (\eg~text, code, image, video), that matches or exceeds the modality-specialized models of similar capacities.}
Our definition aligns with the user experience of proprietary multimodal models such as GPT-4o or Gemini-1.5, where a user does not need to differentiate inputs from different modalities.
Instead, the model is expected to seamlessly handle and integrate multiple modalities' input with a single model.

While proprietary multimodal native models are not uncommon, their training recipes remain largely undisclosed. As a result, most open-source models are modal-specialized or show subpar performance across modalities. In this research, we fill the gap and introduce training recipes for developing multimodal native models from scratch, which includes key aspects below:
\begin{itemize}
    \item \textbf{Model Architecture.} The core of our model is a fine-grained mixture-of-experts decoder,
    which enables faster training and inference speed over dense decoders,
    due to more efficient parameter utilization through expert specialization. 
    \smallname~MoE activates 3.5B parameters per text token and has a total of 24.9B parameters. Visual input of variable length, size, and aspect is encoded as visual tokens using a lightweight visual encoder of 438M parameters. \smallname~has a long multimodal context window of 64k tokens. 
    \item \textbf{Data.} \smallname~is pre-trained on 6.4T language tokens and 400B multimodal tokens. We develop a rigorous procedure to curate high-quality data from a diverse set of sources. The multimodal pre-train data includes four major categories: interleaved image-text sequence from common crawl, synthetic image captions, documents transcriptions and question-answering pairs,  synthetic video captions and question-answering pairs.
    \item \textbf{Training Pipeline.} We design a 4-stage training pipeline, including
    language pre-training, multimodal pre-training, multimodal long-context pre-training, and multimodal post-training.
    Each stage is designed to progressively enhance certain model capabilities while maintaining those acquired in early stages.
    Our pipeline efficiently and effectively exploits the data and compute resources to maximize model performance.
\end{itemize}


Following this recipe, \smallname~demonstrates state-of-the-art performance as an open multimodal native model.
Compared to Pixtral-12B~\citep{pixtral} and Llama3.2-11B~\citep{llama3},
Aria demonstrates superior performance across a wide range of multimodal, language, and coding tasks,
while enjoying lower inference cost due to the fewer number of activated parameters.
In addition, \name~also performs on par with proprietary models such as GPT-4o and Gemini-1.5 on various multimodal tasks.
The detailed benchmark results are present in Table~\ref{table:main}.

\definecolor{LightRed}{rgb}{0.84, 0.84, 1}
\renewcommand{\arraystretch}{1.1} 
\begin{table}[!t]
\centering
\resizebox{\textwidth}{!}{
\begin{tabular}{l|l|>
{\columncolor{LightRed}}c|c|c|c|c|c|c|c}
\hline
\textbf{Category} & \textbf{Benchmark} & \rotatebox{90}{\textbf{\smallname}} & \rotatebox{90}{\textbf{Pixtral-12B}}&  \rotatebox{90}{\textbf{Llama3.2-11B}} &  
\rotatebox{90}{\textbf{GPT-4V}} &
\rotatebox{90}{\textbf{GPT-4o mini}} &
\rotatebox{90}{\textbf{GPT-4o}} &  
\rotatebox{90}{\textbf{Gemini-1.5 Flash~}} &
\rotatebox{90}{\textbf{Gemini-1.5 Pro}} \\ \hline
\multirow{2}{*}{\makecell[l]{\textbf{Knowledge/Math} \\ \textbf{(Multimodal)}}} 
& MMMU (val) & 54.9 & 52.5  & 50.7 & 56.4 & 59.4 & 69.1 & 56.1 & 62.2\\ 
& MathVista (testmini) & 66.1 & 58.0  & 51.5 & - & 54.7 & 63.8 & 58.4 & 63.9\\ \hline
\multirow{3}{*}{\makecell[l]{\textbf{Document/Chart/} \\ \textbf{Scene Text} \\ \textbf{Understanding}}} 
& DocVQA (test) & 92.6 & 90.7  & 88.4 & 88.4 & - & 92.8 & 89.9 & 93.1  \\ 
& ChartQA (test) &86.4 & 81.8  & 83.4 & 78.4 & - & 85.7 & 85.4 & 87.2\\ 
& TextVQA (val) & 81.1 & - & -  & 78.0 & - & - & 78.7 & 78.7 \\ \hline
\textbf{General Visual QA} & MMBench-1.1 & 80.3  & - & - & 79.8 & 76.0 & 82.2 & - & 73.9\\ \hline
\multirow{3}{*}{\makecell[l]{\textbf{Long Video} \\ \textbf{Understanding}}} 
& EgoSchema (test) & 66.8  & - & - & - & - & 72.2 & 65.7 & 72.2\\ 
& LongVideoBench (test) & 65.3 & 47.4  & 45.7 & 60.7 & 58.8 & 66.7 & 62.4 & 64.4 \\ 
& VideoMME (w subs) & 72.1 & 47.5  & 50.2 & 63.3 & 68.9 & 77.2 & 75.0 & 81.3\\\hline
\multirow{2}{*}{\makecell[l]{\textbf{Knowledge/Math/} \\ \textbf{Reasoning} \\ \textbf{(Language)}}} 
& MMLU (5-shot) & 73.3 & 69.2 & 69.4  & 86.4 & - & 89.1 & 78.9 & 85.9\\ 
& MATH (CoT) & 50.8 & 48.1 &  51.9 & - & 70.2 & 76.6 & - & - \\ 
& ARC Challenge & 91.0 & - &  83.4 & - & 96.4 & 96.7 & - & -\\ \hline
\textbf{Coding}
& HumanEval & 73.2 & 72.0 &72.6  & 67.0 & 87.2 & 90.2 & 74.3 & 84.1\\ 
\hline
\end{tabular}
}
\vspace{1ex}
\caption{Performance comparison across various multimodal and language benchmarks. Results of
competing models are collected from verified official sources or reruned with official settings.}
\vspace{-1ex}
\label{table:main}
\end{table}

We release \smallname~under the Apache 2.0 license,
free for both academic and commercial use.
To facilitate easier adoption,
we open-source a training framework that enables finetuning \smallname~on a wide variety of data sources and formats,
using as few as one GPU.



\section{Model}
\label{sec:model}


\subsection{Fine-Grained Mixture-of-Experts}

MoE has emerged as a preferred architecture over dense models for building compute-efficient large language models~\citep{switch_transformer,mixtral,deepseekmoe,moe_scaling_low}.
The core idea of MoE is to replace each feed-forward layer (FFN) in a Transformer with a set of experts, where each expert is structurally identical to an FFN.
Each input token is routed to only a subset of experts in each layer.
The sparsity of expert activation ensures computational efficiency of an MoE layer.

Due to the vast diversity of multimodal data,
we hypothesize that \textit{expert specialization} is important for an multimodal MoE to understand input from different data distributions.
To this end,
we use a large number of fine-grained experts with smaller FFN hidden dimension than standard FFNs, similar to~\citep{deepseekmoe}.
In particular, 
\smallname~has 66 experts in each MoE layer,
2 of the 66 experts are shared among all inputs to capture common knowledge,
whereas 6 more experts are activated for each token by a router module.
Table~\ref{table:moe} shows the detailed architectural configuration.

\smallname~is significantly different from previous multimodal MoEs which either design modality-specific expert architectures or rely on upcycling from dense models~\citep{moma,vl_moe,moe_llava}.
Our multimodal native MoE is pre-trained from scratch with modality-generic experts.
In Section~\ref{sec:expert_analysis},
we show that multimodal expert specialization naturally arises after pre-training.

\begin{table}[h!]
\centering
\resizebox{\textwidth}{!}{
\begin{tabular}{c c c c c c c}
\hline
\textbf{\#total parameters} & \textbf{\#activated parameters} & \textbf{\#experts}& \textbf{\#activated experts} & \textbf{expert FFN dim} & \textbf{hidden dim}& \textbf{\#layers} \\ \hline
24.9B & 3.5B & 2$^{\triangle}+$64 & 2$^{\triangle}+$6 & 1664 & 2560 & 28 \\
\hline
\end{tabular}
}
\caption{Architectural configuration of our MoE decoder. $^{\triangle}$ denotes shared experts.}
\label{table:moe}
\end{table}









\subsection{Visual Encoder}
We design a lightweight visual encoder to convert visual inputs (i.e. images or video frames) into continuous visual tokens with the same feature dimension as word embeddings,
which enables the MoE to seamlessly integrate visual and language inputs.

Drawing inspiration from previous work~\citep{blip2,qwen-vl,idefics2},
our visual encoder consists of a Vision Transformer (ViT) and a projection module.
The ViT accepts images in their native aspect ratio as variable-length sequences of patches~\citep{pix2struct,navit},
which preserves the inherent information structure in images.
We categorize image size into three ranges:
(1) medium-resolution images, where the longer edge is resized to 490 pixels;
(2) high-resolution images, where the longer edge is resized to 980 pixels;
(3) ultra-high-resolution images, where an image is dynamically decomposed into multiple high-res images, following a strategy similar to~\cite{llava1.5}.
We initialize the weights of our ViT using the SigLIP-SO400M model~\citep{siglip} and continue pre-train the ViT on our multimodal data.

Our projection module transforms the sequence of image embeddings from the ViT into a  sequence of visual tokens.
It comprises a single cross-attention layer and a FFN layer.
The cross-attention layer employs a set of trainable vectors as queries and the image embeddings as keys.
Medium-resolution images are processed by 128 queries,
whereas high-resolution images are processed by an additional 128 queries (256 queries in total).
The outputs from the cross-attention layer are then fed to an FFN,
which then outputs visual tokens for the MoE decoder to further process.

\subsection{Infrastructure}

\smallname~is trained on an extensively modified Megatron framework \citep{shoeybi2019megatron}. We eschew pipeline parallelism and instead implement a combination of expert parallelism \citep{lepikhin2020gshard} and ZeRO-1 data parallelism \citep{rajbhandari2020zero} to optimize performance. Due to the carefully designed parallelism method and the small model size, \name~can be effectively trained without using tensor parallelism,
which significantly reduces communication overhead and enhances training efficiency.

We implement a load balancing loss to prevent routing collapse and encourage balanced expert activation.
We find that the expert-level load balancing loss in previous work~\citep{switch_transformer,deepseekmoe} is overly restrictive for our MoE due to the large number of experts.
Therefore, we relax the load balancing to groups of experts, where each group contains 8 fine-grained experts.
We also employ z-loss~\citep{zoph2022st} to stabilize training.

\section{Training}
\label{sec:train}
In this section, we delineate our 4-stage training pipeline.
In each stage, the model aims to learn new capabilities while maintaining those acquired previously. 
We perform evaluation during each stage to ensure that such goal is achieved in a data-efficient and compute-efficient way.

\subsection{Language Pre-training}
The first stage pre-trains the MoE decoder with a large amount of curated language data converted into discrete text tokens,
using a next-token prediction loss,
which enables the MoE to learn general knowledge about the world.
The context window length is 8K tokens.

\noindent{\textbf{Language Data.}}
Our language pre-training data contains 6.4T tokens in total, curated from a variety of data sources containing knowledge until May 2024. 
We de-duplicate the data at different granularities and perform rigorous quality filtering, using a combination of rule-based approach and model-based quality classifiers.
To enhance model's in-context learning capability,
we employ data clustering and pack similar data in the same sequence during training, akin to the approach in \cite{shi2023incontext}. 
However, their original method is less scalable and likely to generate numerous long-tail structures when processing trillions of tokens. Instead, we utilize a minimum spanning tree algorithm for language data clustering, which resulted in a noticeable performance gain.

\subsection{Multimodal Pre-training}
The second stage pre-trains the MoE decoder and the visual encoder with a mixture of language and multimodal data, using the same next-token prediction loss.
This stage aims to enable the model with broad multimodal understanding abilities,
while maintaining or even improving its language understanding.
To this end, the language data contains a high-quality subset of 1T tokens, covering topics including code, reasoning, and knowledge.
The multimodal data contains 400B tokens from a diverse set of sources, which can be categorized into four major categories below.

\noindent{\textbf{Interleaved image-text web data.}}
We extract and filter web pages from Common Crawl.
The filtering process first removes web pages with low image or text quality.
Then, it de-duplicate images, and removes web pages where the images and the text have low overall CLIP score~\citep{clip}.
Additionally, we adjust the position of the images in the sequence,
by moving an image to the front of a sentence if the sentence has higher CLIP score and is in front of the image.
In total, we curate 190B interleaved image-text tokens.

\noindent{\textbf{Synthetic image captions.}}
Alt texts directly extracted for web images are generally short, less descriptive, and noisy.
It has been shown in previous work that synthetic data at scale can improve multimodal pre-training~\citep{blip}.
We thus synthesize image captions using a small model which has learned to generate longer and more descriptive image captions by re-writing the alt texts.
We create synthetic captions for 300M images in the LAION-400M dataset~\cite{laion},
resulting in a total of 70B multimodal tokens.

\noindent{\textbf{Document transcriptions and QA.}}
To improve the model's capability of understanding text-heavy images,
we transcribe document images into texts using public OCR methods.
We also render images using plain text, chart json or table/equation latex code.
In order to enhance the model's ability to not only transcribe text but also understand its meaning,
we use a language model to create synthetic question-answering pairs.
In total, our multimodal document data contains 102B tokens.

\noindent{\textbf{Video captions and QA.}}
We collect 4.4M videos of varying lengths from a diverse range of sources.
We train a model to generate frame-level dense descriptions for the videos.
Then, we use a language model to generate question-answering pairs and video summarizations based on the dense video descriptions.
In total, our video data contains 35B tokens.
We select samples within 8K length for multimodal pre-training.

\vspace{-2ex}
\subsection{Multimodal Long-Context Pre-training}
In this stage, we pre-train on long sequences to extend the model's context window to 64K tokens.
Language long-sequence data is selected from the pre-train data source.
Multimodal long-sequence data contains long videos, long documents and synthetic long sequences constructed from short multimodal data.
In particular, we concatenate a sequence of independent images as input, and concatenate their image descriptions as target.
This stage consumes 12B language tokens and 21B multimodal tokens, 
where 69\% of the 33B tokens are long sequences.
We increase the RoPE base frequency hyperparameter from 100K to 5M.

After this stage, the model perfectly solves the needle-in-a-haystack task~\citep{niah} for up to 64K context window.
It also demonstrates substantial performance improvement on long video understanding and long multimodal document understanding tasks.

\subsection{Multimodal Post-training}
The final post-training stage anneals the learning rate to converge the model.
The learning focuses on improving the model's question-answering and instruction-following capabilities,
using a mixture of high-quality open-source datasets and human-annotated datasets,
covering domains including multimodal, code, math, and reasoning.
This stage digests 20B tokens in total.






\section{Evaluation and Analysis}
\subsection{Benchmark Results}

\definecolor{LightRed}{rgb}{0.84, 0.84, 1}
\renewcommand{\arraystretch}{1.14} 
\setlength{\tabcolsep}{10pt} 
\begin{table}[!t]
\centering
\resizebox{\textwidth}{!}{
\begin{tabular}{l c |c c|c c|c c}
\hline
\multirow{2}{*}{\textbf{Model}} & \textbf{\#Params} & \multicolumn{2}{c|}{\textbf{LongVideoBench}} & \multicolumn{2}{c|}{\textbf{VideoMME}} & \multicolumn{2}{c}{\textbf{MMLongBench-Doc}} \\
 &  activated (total)&test & val & w subs & w/o subs & acc & f1 \\
 \hline
 \multicolumn{2}{c|}{\textit{Open-source}}& & & & & & \\
\rowcolor{LightRed} \textbf{\name} & 3.9B (25.3B) & 65.3 & 63.0 & 72.1 & 67.6 & 28.3 & 24.6 \\ 
\textbf{Qwen2-VL-7B} & 7B & 56.8 & 55.6 & 69.0 & 63.3 & 21.3 & 22.7 \\
\textbf{Idefics2} & 8B & 49.4 & 49.7 &- & - &7.0&6.8 \\
\textbf{MiniCPM-V-2.6} & 8B  & 55.7 & 54.9 & 63.7 & 60.9 & 11.5 & 11.6 \\
\textbf{Llama3.2-11B} & 11B & 45.7 & 45.5 & 49.5 & 46.0 & 13.8 & 11.3 \\
\textbf{Pixtral-12B} & 12B & 47.4 & 44.9 & 47.5 & 40.7 & 6.4 & 6.0 \\
\textbf{InternVL-Chat-V1.5} & 26B & 51.7 & 51.2 &52.4 & 50.7 & 14.6 & 13.0 \\
\textbf{InternVL2-40B} & 40B & 60.6 & 59.3 & 62.4 & 61.2 & 18.2 & 17.9  \\
\textbf{LLaVA-OneVision-72B} & 72B & 63.2 & 61.3 & 69.6 & 66.3 &- & - \\
\textbf{Qwen2-VL-72B} & 72B & 61.7 & 60.4 & 77.8 & 71.2 & 33.3 & 35.7 \\
\hline
 \multicolumn{2}{c|}{\textit{Proprietary}}& & & & & & \\
\textbf{Gemini-1.5-Flash} & - & 62.6 & 61.4 & 75.0 & 60.3 & 27.0 & 21.3 \\
\textbf{Gemini-1.5-Pro} & - & 64.4 & 64.0 & 81.3 & 75.0 & 28.2 & 20.6 \\
\textbf{GPT-4o mini} & - & 58.8 & 56.5 & 68.9 & 64.8 & 29.0 & 28.6 \\
\textbf{GPT-4o}  & - & 66.7 & 66.7 & 77.2 & 71.9 & 42.9 & 44.9 \\
\hline
\end{tabular}
}
\vspace{1ex}
\caption{Evaluation of long-context multimodal understanding on videos and documents. Results of competing models are collected from verified official leaderboards or reruned with official settings.}
\label{table:longcontext}
\vspace{-1ex}
\end{table}

In Table~\ref{table:main},
we compare \name~against leading open models of similar scale and proprietary models across a variety of established benchmarks.
In Table~\ref{table:longcontext} and Table~\ref{table:instruction_following},
we examine the long-context multimodal understanding and instruction following capability, respectively.
Based on the evaluation result, we highlight the following key observations.

\noindent\textbf{\name~is the best-in-class open multimodal native model},
showing clear advantages over Pixtral-12B and Llama3.2-11B across a wide range of multimodal, language, and coding tasks.

\noindent\textbf{\name~is competitive against proprietary models on various multimodal tasks}, including document understanding, chart reading, scene text recognition, and video understanding.

\noindent\textbf{\name~excels in long-context multimodal understanding.}
Real-world multimodal data is complex by nature and often involves long sequences of interleaved vision-language input, such as videos with subtitles or multi-page documents. 
\name~excels in understanding such data, significantly outperforming open models such as Qwen2-VL-7B~\citep{qwen-vl} and LLaVA-OneVision-72B~\citep{llava_ov}. Compared to proprietary models, \name~outperforms GPT-4o mini in long video understanding~\citep{wu2024longvideobench}, and is superior to Gemini-1.5-Flash in long document understanding, making \name~a preferable choice for processing long multimodal data in a compute-efficient and time-efficient manner.

\noindent\textbf{\name~has strong instruction following capabilities},
outperforming other open models on both multimodal and language-only benchmarks. 
See Section~\ref{sec:examples} for qualitative examples.

 \renewcommand{\arraystretch}{1.14}
 \setlength{\tabcolsep}{4pt} 

\begin{table}[!t]
\centering
\resizebox{\textwidth}{!}{
\begin{tabular}{l | >{\columncolor{LightRed}}c c c c c c c c c c c c c}
\hline
 & \rotatebox{90}{\textbf{\name}}& \rotatebox{90}{\textbf{Phi-3 Vision}} & \rotatebox{90}{\textbf{Qwen2-VL-7B}}  & \rotatebox{90}{\textbf{Idefics2}} & \rotatebox{90}{\textbf{Pixtral-12B}} & \rotatebox{90}{\textbf{InternVL-Chat-v1.5}~} & \rotatebox{90}{\textbf{LLaVA-NeXT-34B}} & \rotatebox{90}{\textbf{MiniCPM-V-2.5}} & \rotatebox{90}{\textbf{Gemini-1.0-Pro}} & \rotatebox{90}{\textbf{Reka-Core}} & \rotatebox{90}{\textbf{Claude-3-Sonnet}} & \rotatebox{90}{\textbf{GPT-4o}} \\
\hline
\textbf{MIA-Bench} (Multimodal) & 8.76 & 7.60  & 8.07 & 5.14 & 8.43  & 7.54  & 7.56 & 7.63 & 7.06 & 7.70 & 7.94  & 8.86 \\
\textbf{MT-Bench} (Language) & 8.53 & 6.27& 6.41 & - & 7.68 & - & -  & - & - & - & - & - \\ 

\hline
\end{tabular}
}
\vspace{1ex}
\caption{Evaluation of instruction following capabilities. Results of competing models are copied from~\cite{miabench} for MIA-Bench and~\cite{pixtral} for MT-Bench.}
\label{table:instruction_following}
\vspace{-1ex}
\end{table}

\subsection{Expert Modality Specialization}
\label{sec:expert_analysis}

We analyze the expert activation behavior across all layers in \name~MoE after the multimodal pre-training stage.
We use multimodal data from three domains for analysis: natural image, video, and PDF-format image.
For each expert, we first compute its activation rate for both visual tokens and text tokens, denoted as $R_v$ and $R_t$.
$R_v$ refers to the number of visual tokens that activates the expert divided by the total number of visual tokens processed by all experts of that layer,
and $R_t$ refers to the same meaning for text tokens.
Then we compute the ratio $R_v/R_t$,
which represents the expert's level of visual specialization.
A higher visual specialization suggests that the expert is more frequently activated by visual tokens compared to text tokens.

Figure~\ref{fig:expert_routing} shows a visualization of the expert visual specialization value (capped at 50).
It is observed that a number of visual-specialized experts exist in most layers.
Furthermore, multiple layers (layer 4, 5, 14, 15, 16, 17, 20) have a single visual expert specialized in all three visual domains.
This analysis shows that despite its modality-generic architecture,
our MoE learns effective expert utilization during pre-training.
\begin{figure}[H]
\centering
\foreach \row in {0,...,2} {%
    \foreach \col in {0,1} {%
        \pgfmathtruncatemacro{\imageindex}{(\row)*2 + \col}%
        \includegraphics[width=0.49\textwidth]{moe_token_figures/heatmap_layer_\imageindex_ckpt_43000.png}%
        \ifnum\col=0\hfill\fi%
    }%
    \par
}%
\end{figure}

\begin{figure}[H]
\centering
\foreach \row in {3,...,13} {%
    \foreach \col in {0,1} {%
        \pgfmathtruncatemacro{\imageindex}{(\row)*2 + \col}%
        \includegraphics[width=0.49\textwidth]{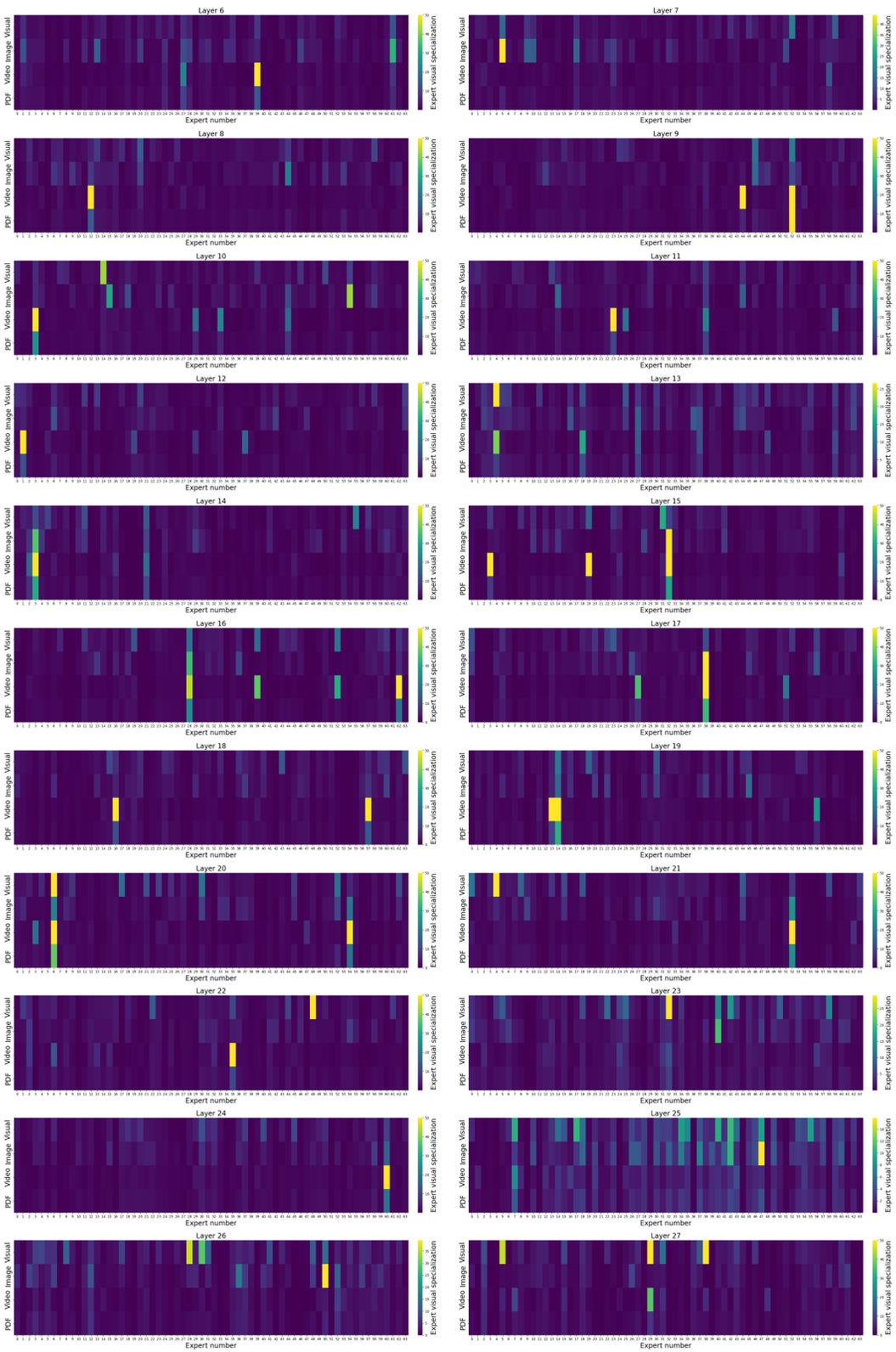}%
        \ifnum\col=0\hfill\fi%
    }%
    \ifnum\row<13\par\fi%
}%
\caption{Visualization of expert specialization in each MoE layer.}
\label{fig:expert_routing}
\end{figure}

\subsection{Qualitative Analysis}
\label{sec:examples}
\subsubsection{Multimodal Native Reasoning with Vision, Language, Coding Capabilities}

\begin{tabular}{l p{12cm}}
    \toprule
    \multicolumn{2}{c}{\textbf{\name~Example - Weather Forecast Information Extraction}}\\ 
    \midrule
    \textbf{User} &  Organize the highest temperature, lowest temperature, and humidity for all even dates into a table.

Write a python code to draw a line chart for the temperatures, and a bar chart for the humidity. Use red and green colors for the temperatures, and use blue color for the humidity.\\
& \begin{center}
\includegraphics[width=0.65\textwidth]{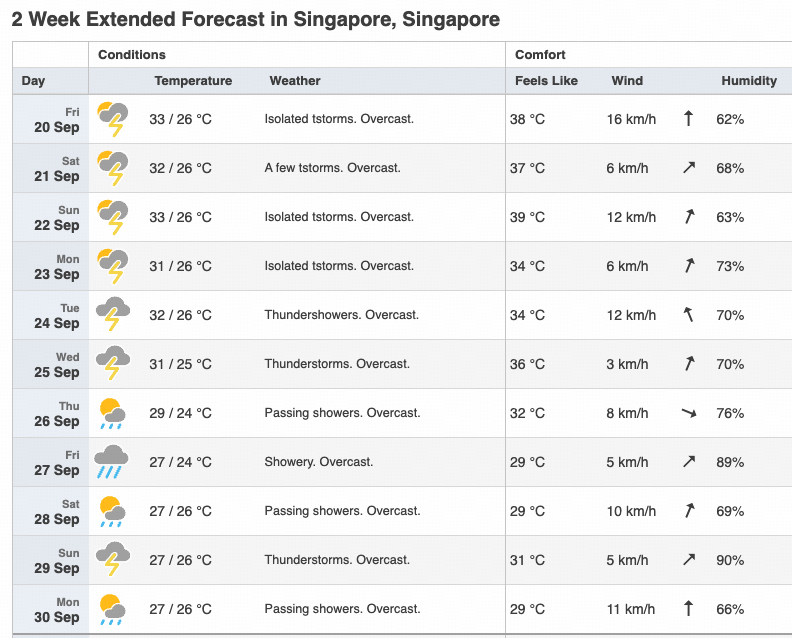}    
\end{center}
 \\
\midrule
\textbf{\name} & {Table of values for even dates}:
\begin{center}
\includegraphics[width=0.6\textwidth]{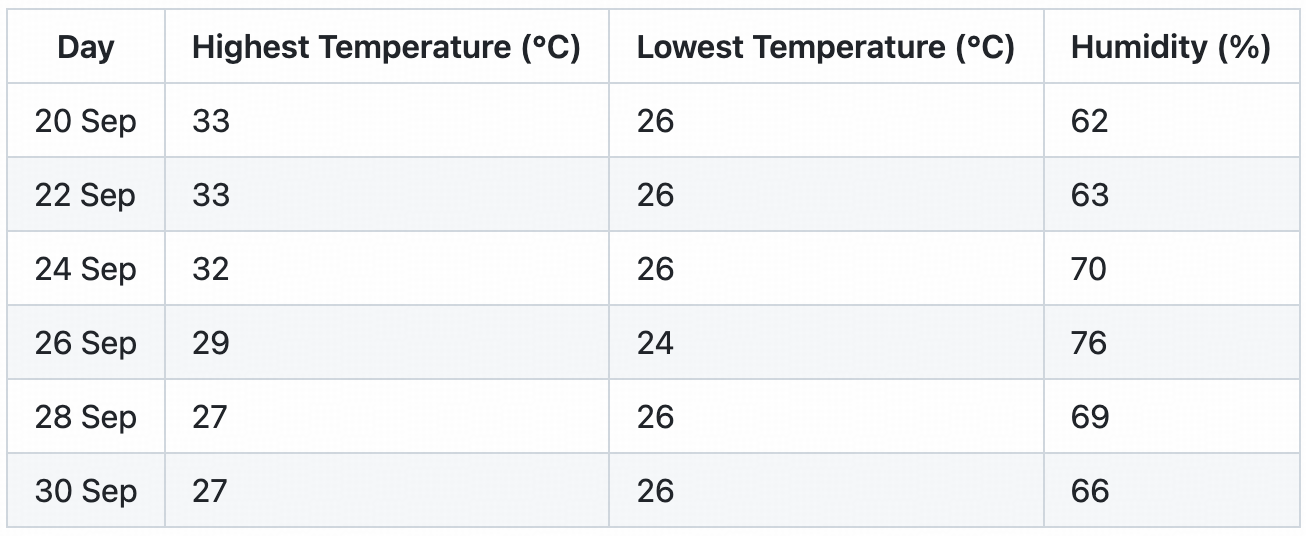}    
\end{center}
\begin{lstlisting}[mathescape=true]
import matplotlib.pyplot as plt

# Data for even dates
days = ['20 Sep', '22 Sep', '24 Sep', '26 Sep', '28 Sep', '30 Sep']
high_temps = [33, 33, 32, 29, 27, 27]
low_temps = [26, 26, 26, 24, 26, 26]
humidity = [62, 63, 70, 76, 69, 66]

# Line chart for temperatures
plt.figure(figsize=(14, 6))

plt.subplot(1, 2, 1)
plt.plot(days, high_temps, 'r-', label='Highest Temperature', marker='o')
plt.plot(days, low_temps, 'g-', label='Lowest Temperature', marker='o')
plt.xlabel('Days')
plt.ylabel('Temperature ($\color{deepred} ^\circ$C)')

plt.title('Temperature Variation')
plt.legend()
plt.grid(True)
\end{lstlisting}

\\
    \bottomrule
\end{tabular}
\newpage
\begin{tabular}{l p{12cm}}
    \toprule
    \multicolumn{2}{c}{\textbf{\name~Example - Weather Forecast Information Extraction}}\\
\midrule
\textbf{\name} &(continued)
\begin{lstlisting}
# Bar chart for humidity
plt.subplot(1, 2, 2)
plt.bar(days, humidity, color='blue', alpha=0.7)
plt.xlabel('Days')
plt.ylabel('Humidity (%)')
plt.title('Humidity')
plt.ylim(0, 100)
plt.grid(axis='y')

plt.tight_layout()
plt.show()
\end{lstlisting}
\\
\midrule
\textbf{Result} & Figures generated by the code provided by \name~is shown as below.
\begin{center}
\includegraphics[width=0.8\textwidth]{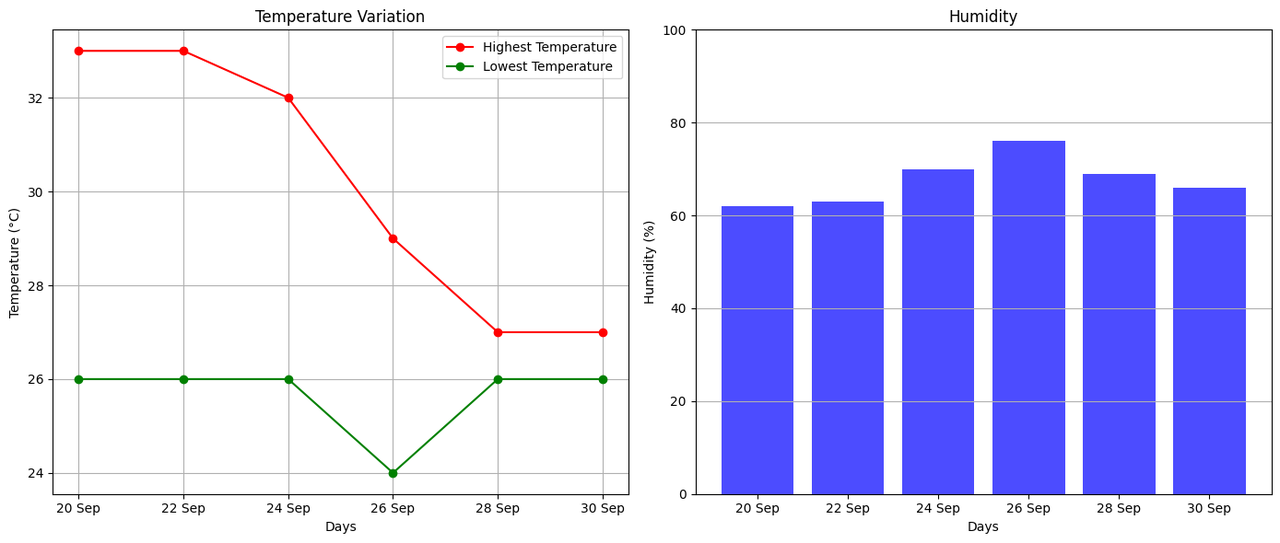}
\end{center}\\
\bottomrule
\end{tabular}
\footnote{Weather forecast image source (14-day forecast, screenshot on Sep 20, 2024): \\ \url{https://www.timeanddate.com/weather/singapore/singapore/hourly}}

\newpage
\begin{tabular}{l p{12cm}}
    \toprule
    \multicolumn{2}{c}{\textbf{\name~Example - Financial Report Analysis}}\\ 
    \midrule
    \textbf{User} &  How does the company's profit margin change from year 2020 to 2021? What factors cause such changes? Answer step by step.\\
& \begin{center}
\includegraphics[width=0.7\textwidth]{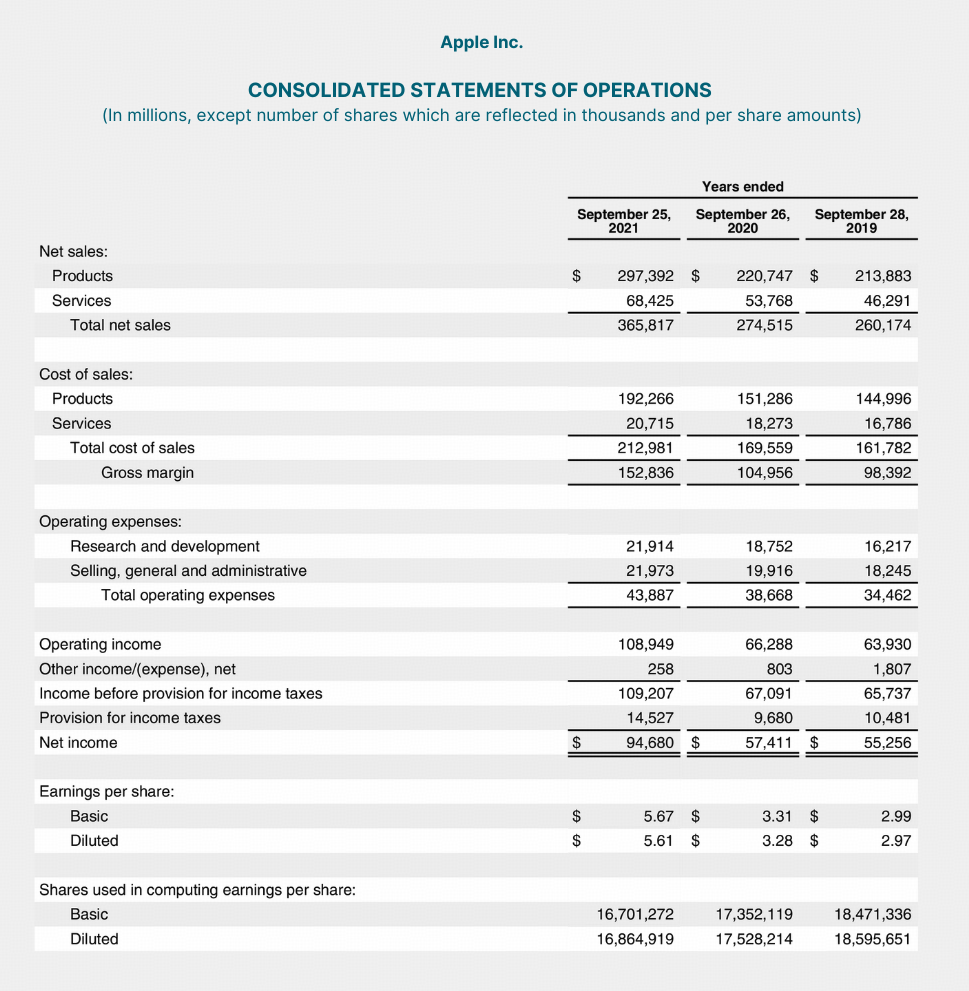}    
\end{center}
 \\
\midrule
\textbf{\name} & 
To determine how the company's profit margin changed from 2020 to 2021, we need to follow these steps:

\begin{enumerate}[leftmargin=*]
    \item \textbf{Calculate the profit margins for both years:}
    \begin{itemize}[leftmargin=*]
        \item \textbf{2020:}
        \begin{itemize}
            \item Net income: \$57,411 million
            \item Total net sales: \$274,515 million
            \item Profit margin:
            \[
            \text{Profit Margin} = \left( \frac{\text{Net Income}}{\text{Total Net Sales}} \right) \times 100 = \left( \frac{\$57,411}{\$274,515} \right) \times 100 \approx 20.91\%
            \]
        \end{itemize}
        \item \textbf{2021:}
        \begin{itemize}
            \item Net income: \$94,680 million
            \item Total net sales: \$365,817 million
            \item Profit margin:
            \[
            \text{Profit Margin} = \left( \frac{\text{Net Income}}{\text{Total Net Sales}} \right) \times 100 = \left( \frac{\$94,680}{\$365,817} \right) \times 100 \approx 25.88\%
            \]
        \end{itemize}
    \end{itemize}

\end{enumerate}
\\
\bottomrule
\end{tabular}

\begin{tabular}{l p{12cm}}
    \toprule
    \multicolumn{2}{c} {\textbf{\name~Example - Financial Report Analysis}} \\
    \midrule
\textbf{\name} & (continued) \begin{enumerate}[leftmargin=*]
  \setcounter{enumi}{1}
    \item \textbf{Analyze the factors causing the change:}
    \begin{itemize}
        \item \textbf{Increase in Net Income:} The net income increased from \$57,411 million in 2020 to \$94,680 million in 2021, showing a significant rise of \$37,269 million.
        \item \textbf{Sales Growth:} Total net sales also increased from \$274,515 million in 2020 to \$365,817 million in 2021, indicating a strong growth in revenue.
        \item \textbf{Cost Management:} The cost of sales increased from \$151,286 million in 2020 to \$192,266 million in 2021, but the percentage increase in sales outpaced the increase in costs, leading to a higher profit margin.
        \item \textbf{Operating Expenses:} Operating expenses grew from \$38,668 million in 2020 to \$43,887 million in 2021, but the rise in sales was more substantial, positively impacting the profit margin.
    \end{itemize}
\end{enumerate}

Therefore, the company's profit margin improved from approximately 20.91\% in 2020 to approximately 25.88\% in 2021. This improvement is primarily due to a substantial increase in net income and effective cost management relative to the growth in total net sales.
\\
\bottomrule
\end{tabular}
\footnote{Financial report image source: \\ \url{https://www.financestrategists.com/wealth-management/financial-statements/}}
    

\newpage
\begin{tabular}{l p{12cm}}
    \toprule
    \multicolumn{2}{c}{\textbf{\name~Example - Handwritten Math Equation Explanation}}\\ 
    \midrule
    \textbf{User} &  Explain the equation according to the notes, then give an example with numbers.\\
& \begin{center}
\includegraphics[width=0.7\textwidth]{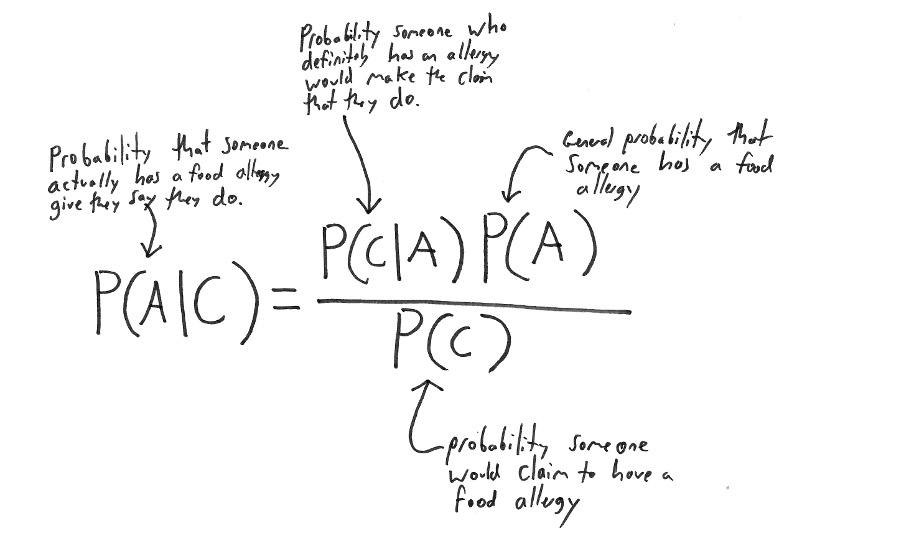}    
\end{center}
 \\
\midrule
\textbf{\name} & The equation in the image represents Bayes' Theorem, which is used to update the probability estimate of a hypothesis (in this case, having a food allergy) given new evidence (claiming to have a food allergy).
\newline\newline
\textbf{Explanation:}
\begin{itemize}
    \item $P(C|A)$: Probability that someone who definitely has an allergy would claim that they do.
    \item $P(A)$: General probability that someone has a food allergy.
    \item $P(C)$: Probability that someone would claim to have a food allergy.
    \item $P(A|C)$: Probability that someone actually has a food allergy given that they claim to have one.
\end{itemize}

The equation:
\[
P(A|C) = \frac{P(C|A) \cdot P(A)}{P(C)}
\]

\textbf{Example with numbers:}

Assume:
\begin{itemize}
    \item $P(C|A) = 0.9$, \text{(90\%\, of people with an allergy claim they do)}
    \item $P(A)   = 0.05$, \text{(5\%\, of the population has an allergy)}
    \item $P(C)   = 0.1$, \text{(10\%\, of the population claims to have an allergy)}
\end{itemize}

Using the formula:
\[
P(A|C) = \frac{0.9 \times 0.05}{0.1} = 0.45
\]

So, the probability that someone who claims to have a food allergy actually does is 45\%.

\\
    \bottomrule
\end{tabular}
\footnote{Image source: \\ \url{https://ngathanblog.wordpress.com/2020/06/25/baysian-struggle/}.}
\newpage
\begin{tabular}{l p{12cm}}
    \toprule
    \multicolumn{2}{c}{\textbf{\name~Example - Code Debugging from Screenshot}}\\ 
    \midrule
    \textbf{User} &  Debug this code.\\
& \begin{center}
\includegraphics[width=0.7\textwidth]{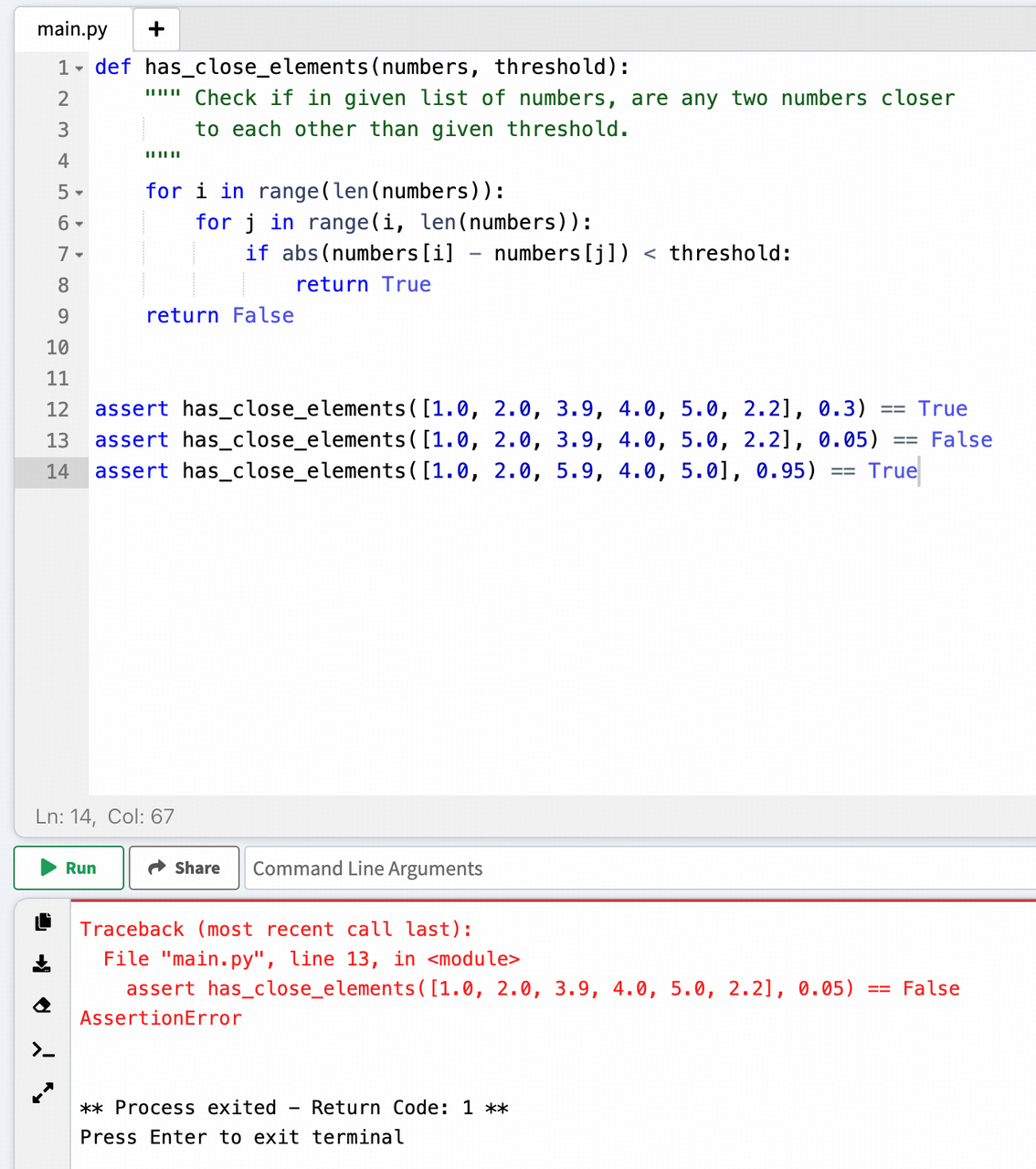}    
\end{center}
 \\
\midrule
\textbf{\name} & The code has a logical error in the nested loop logic. The current implementation of the \texttt{has\_close\_elements} function does not correctly check if any two numbers in the list are closer to each other than the given threshold.

Here's the issue:
\begin{enumerate}
    \item The current condition \texttt{if abs(numbers[i] - numbers[j]) < threshold:} checks the absolute difference between each pair of numbers but should only check pairs where \texttt{i} and \texttt{j} are different indices.
    \item The inner loop should start from \texttt{i + 1} instead of \texttt{i} to avoid comparing a number with itself.
\end{enumerate}

Here's the corrected version:
\begin{lstlisting}
def has_close_elements(numbers, threshold):
    """
    Check if in given list of numbers, are any two numbers closer to 
    each other than the given threshold.
    """
    for i in range(len(numbers)):
        for j in range(i + 1, len(numbers)):  # Start from i + 1 to avoid comparing with itself
            if abs(numbers[i] - numbers[j]) < threshold:
                return True
    return False

# Tests
assert has_close_elements([1.0, 2.0, 3.9, 4.0, 5.0, 2.2], 0.3) == True
assert has_close_elements([1.0, 2.0, 3.9, 4.0, 5.0, 2.2], 0.05) == False
assert has_close_elements([1.0, 2.0, 5.9, 4.0, 5.0], 0.95) == True
\end{lstlisting}
This change ensures the function correctly checks all pairs of numbers and returns \texttt{True} if any pair is closer than the threshold, otherwise \texttt{False}.
\\ 
    \bottomrule
\end{tabular}
\newpage

\begin{tabular}{l p{12cm}}
    \toprule
    \multicolumn{2}{c} {\textbf{\name~Example - Multi-image Food Chain Identification}} \\
    \midrule
    \textbf{User} &  Based on the five species in the images, draw a food chain. Explain the role of each species in the food chain. \\
& \begin{center}
\includegraphics[width=0.8\textwidth]{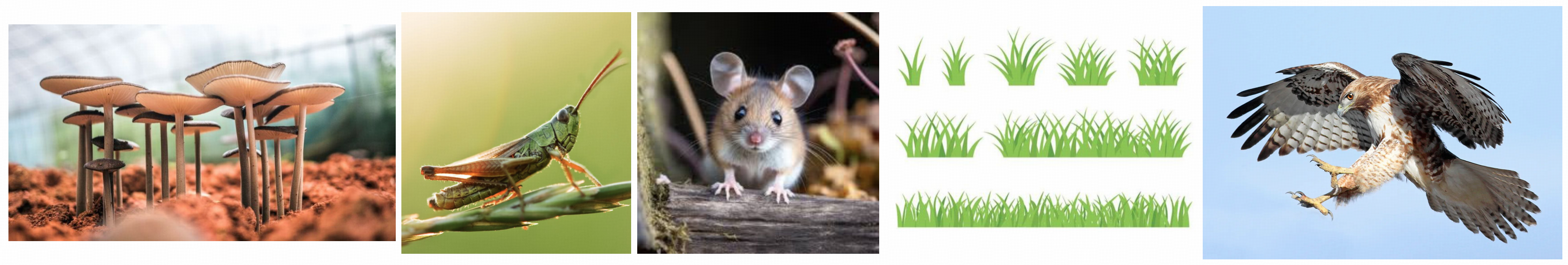}
\end{center}\\
\textbf{\name} & \begin{enumerate}
    \item \textbf{Mushrooms} (\textit{Image 1}): Decomposers, breaking down dead organic matter and recycling nutrients.
    \item \textbf{Grass} (\textit{Image 4}): Primary producers, converting sunlight into energy via photosynthesis.
    \item \textbf{Grasshopper} (\textit{Image 2}): Primary consumers, herbivores feeding on plants.
    \item \textbf{Mouse} (\textit{Image 3}): Secondary consumers, herbivores that eat plants and possibly small insects.
    \item \textbf{Hawk} (\textit{Image 5}): Tertiary consumers, predators that eat mice and other small animals.
\end{enumerate}

\textbf{Food Chain}

\[
\text{Grass} \rightarrow \text{Grasshopper} \rightarrow \text{Mouse} \rightarrow \text{Hawk} \rightarrow \text{Mushrooms}
\]

\textbf{Explanation}
\begin{enumerate}
    \item Grass serves as the base, providing energy to grasshoppers.
    \item Grasshoppers eat the grass and are then consumed by mice.
    \item Mice are preyed upon by hawks.
    \item Hawks are apex predators, feeding on mice.
    \item Mushrooms decompose dead matter, recycling nutrients back into the ecosystem.
\end{enumerate}
\\
\bottomrule
\end{tabular}

\subsubsection{Long Multimodal Context Window}

\begin{tabular}{l p{12cm}}
    \toprule
    \multicolumn{2}{c} {\textbf{\name~Example - Paper Reading}} \\
    \midrule
    \textbf{User} &  Summarize the key contribution of the paper, and explain how it uses various figures and tables to showcase such contribution. \\
& \begin{center}
\includegraphics[width=0.8\textwidth]{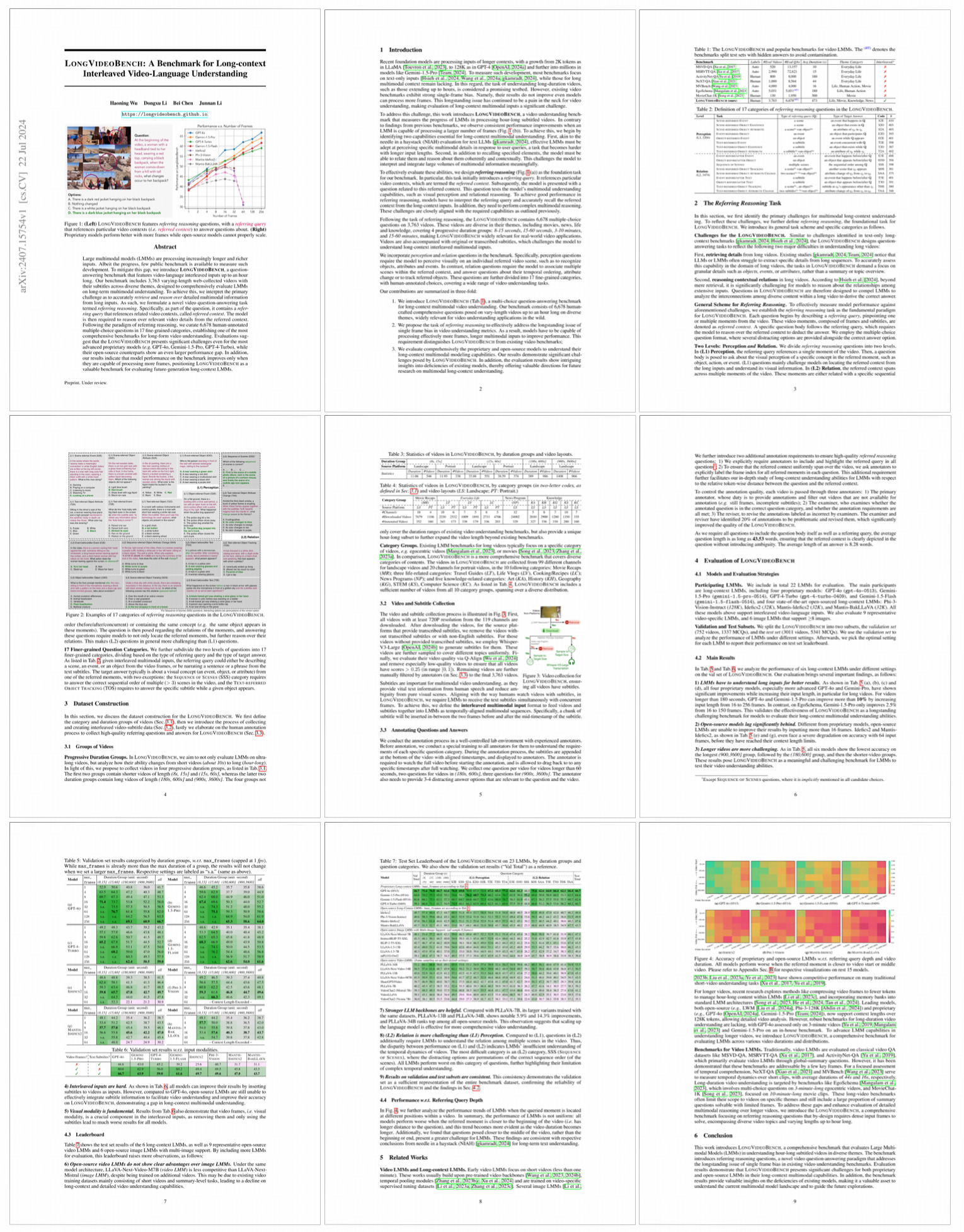}
\end{center}\\
\textbf{\name} &
The paper introduces LONGVIDEOBENCH, a benchmark for evaluating Large Multimodal Models (LMMs) in long-context video-language understanding. It addresses the challenge of processing longer video inputs by proposing a question-answering benchmark that tests the models' ability to understand and reason about video contexts interleaved with up to an hour-long video. This benchmark includes 3,763 videos with diverse themes and 6,678 human-annotated multiple-choice questions across 17 categories.
\newline\newline
The key contribution is the comprehensive evaluation of LMMs' performance on long-form video understanding, highlighting the limitations of existing models in handling extended inputs. The paper demonstrates that while proprietary models like GPT-4 and Gemini-1.5-Pro show improvements with longer inputs, open-source models lag significantly behind.
\\
\bottomrule
\end{tabular}

\begin{tabular}{l p{12cm}}
    \toprule
    \multicolumn{2}{c} {\textbf{\name~Example - Paper Reading}} \\
    \midrule
\textbf{\name} &
(continued) Figures and tables are used to illustrate these points:
\begin{itemize}
    \item \textbf{Figure 1}: Demonstrates the performance of different models on the benchmark, showing how accuracy varies with the number of frames.
    \item \textbf{Figure 2}: Provides examples of the 17 categories of referring reasoning questions.
    \item \textbf{Table 1}: Lists the benchmark and previous benchmarks for comparison.
    \item \textbf{Table 2}: Details the 17 categories of referring reasoning questions.
    \item \textbf{Table 3}: Shows statistics of videos in \textsc{LONGVIDEOBENCH} by category groups and duration.
    \item \textbf{Table 5}: Validation set results categorized by duration groups.
    \item \textbf{Table 7}: Test set leaderboard of LMMs by duration groups and question categories.
\end{itemize}

These visual aids help in understanding the performance metrics, the diversity of the dataset, and the specific challenges posed by longer videos and complex questions, thus emphasizing the need for improved models in long-form video understanding.
\\
\midrule
\textbf{User} &  According to the paper, what are the two major difficulties in understanding long videos? \\
\midrule
\textbf{\name} & The two major difficulties in understanding long videos, as outlined in the paper, are:

 \begin{enumerate}
    \item \textbf{Retrieving details from long videos}: Existing Large Multimodal Models (LMMs) often struggle to extract specific details from long sequences. To accurately assess tasks in LONGVIDEOBENCH, there is a need for models to focus on granular details such as objects, events, or attributes, rather than providing a summary or topic overview.
    \item \textbf{Reasoning contextual relations in long videos}: Questions in LONGVIDEOBENCH require models to analyze the interconnections among diverse contents. This involves understanding the relationships among objects, events, or attributes within the video, which is significantly challenging for extensive inputs. The tasks demand models to derive the correct answer by examining the context and relations across multiple moments in the video.
    \end{enumerate} 
\\
\bottomrule
\end{tabular}
\footnote{LongVideoBench paper source: \url{https://arxiv.org/pdf/2407.15754v1.pdf}.}

\pagebreak

\begin{tabular}{l p{12cm}}
    \toprule
    \multicolumn{2}{c}{\textbf{\name~Example - Long Video Understanding (Scene Splitting)}}\\ 
    \midrule
    \textbf{User} & Please split this video into scenes, providing start time, end time, a title and detailed descriptions for each scene.\\
& \begin{center}
\includegraphics[width=0.8\textwidth]{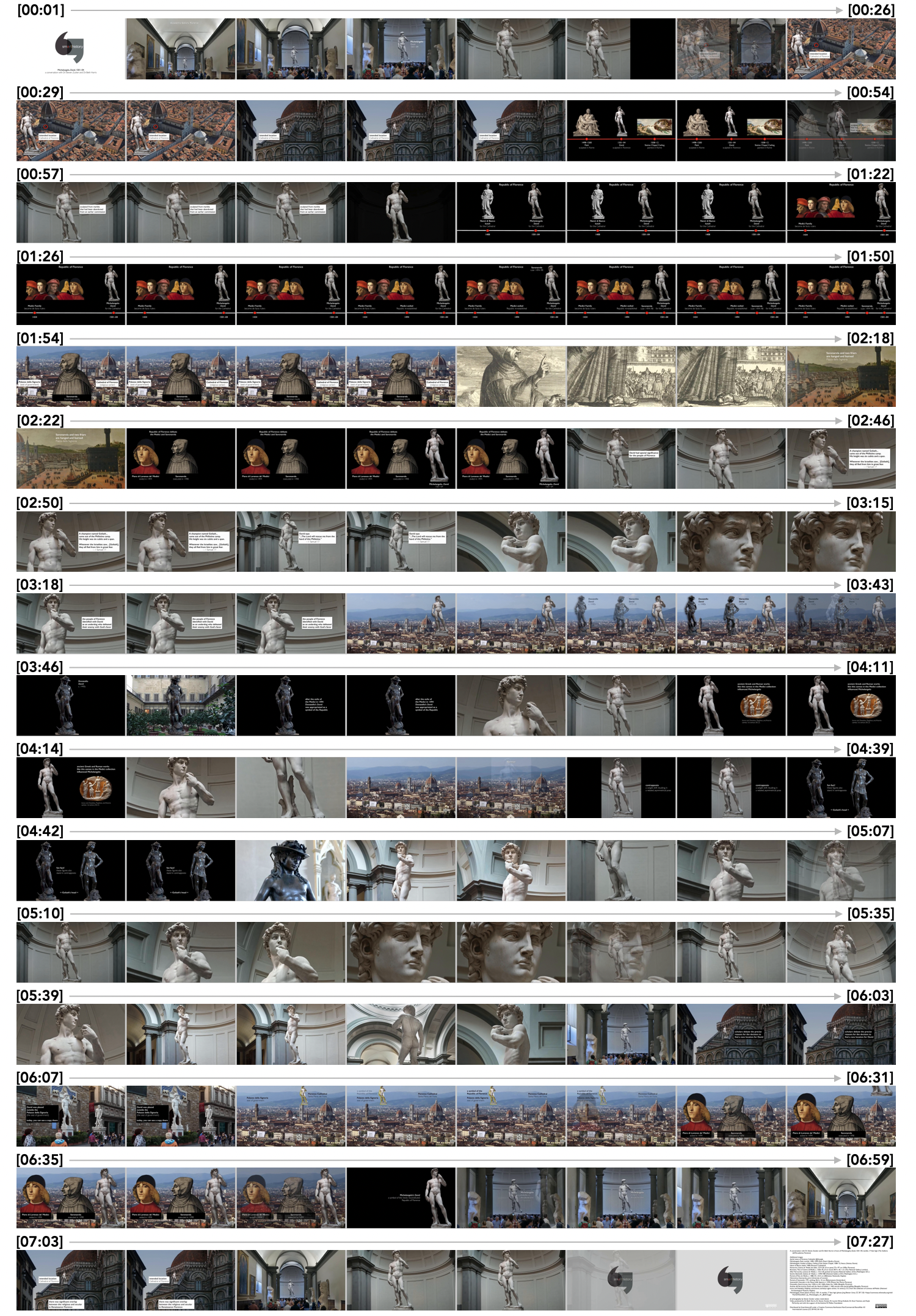}    
\end{center}
\\
\midrule
\textbf{\name} & Certainly! Here are the scenes from the video, along with their start and end times, and detailed descriptions:

\begin{enumerate}
      \item Scene 1: Introduction
        \begin{description}
            \item[Start Time:] 00:01
            \item[End Time:] 00:05
            \item[Title:] Introduction
            \item[Description:] The video opens with a title screen featuring the \textit{smarthistory} logo and the text "Michelangelo, David, 1501-04," indicating a conversation with Dr. Steven Zucker and Dr. Beth Harris.
        \end{description}

\end{enumerate}
\\
\bottomrule
\end{tabular}

\begin{tabular}{l p{12cm}}
    \toprule
    \multicolumn{2}{c} {\textbf{\name~Example - Long Video Understanding (Scene Splitting)}} \\
    \midrule
\textbf{\name} &
(continued) 

\begin{enumerate}[start=2]
    
    \item Scene 2: Accademia Gallery, Florence
        \begin{description}
            \item[Start Time:] 00:05
            \item[End Time:] 00:40
            \item[Title:] Accademia Gallery, Florence
            \item[Description:] The scene transitions to the interior of the Accademia Gallery in Florence, showcasing Michelangelo's David statue surrounded by visitors. The focus is on the statue and its historical context.
        \end{description}

    \item Scene 3: Historical Context
        \begin{description}
            \item[Start Time:] 00:40
            \item[End Time:] 01:08
            \item[Title:] Historical Context
            \item[Description:] The scene shifts to a timeline highlighting significant artworks and events from 1498 to 1512, including Michelangelo's David and the Sistine Chapel ceiling.
        \end{description}

    \item Scene 4: Comparison of Statues
        \begin{description}
            \item[Start Time:] 01:08
            \item[End Time:] 01:50
            \item[Title:] Comparison of Statues
            \item[Description:] The video presents a comparison between Michelangelo's David and Nanni di Banco's Isaiah for the Cathedral, emphasizing their historical significance.
        \end{description}

    \item Scene 5: Republic of Florence
        \begin{description}
            \item[Start Time:] 01:50
            \item[End Time:] 02:08
            \item[Title:] Republic of Florence
            \item[Description:] The scene highlights the Republic of Florence timeline, focusing on the Medici family and Savonarola, with a focus on their impact on the city.
        \end{description}

    \item Scene 6: Savonarola's Execution
        \begin{description}
            \item[Start Time:] 02:08
            \item[End Time:] 02:22
            \item[Title:] Savonarola's Execution
            \item[Description:] The scene depicts the execution of Savonarola and two friars in Piazza della Signoria, emphasizing the historical context.
        \end{description}

    \item Scene 7: Republic Defeats Medici and Savonarola
        \begin{description}
            \item[Start Time:] 02:22
            \item[End Time:] 02:36
            \item[Title:] Republic Defeats Medici and Savonarola
            \item[Description:] The scene shows the Republic of Florence defeating the Medici and Savonarola, with a focus on the historical figures involved.
        \end{description}
\end{enumerate}

\\
\bottomrule
\end{tabular}

\begin{tabular}{l p{12cm}}
    \toprule
    \multicolumn{2}{c} {\textbf{\name~Example, Long Video Understanding (Scene Splitting)}} \\
    \midrule
\textbf{\name} &
(continued) 

\begin{enumerate}[start=8]

    \item Scene 8: David's Significance
        \begin{description}
            \item[Start Time:] 02:36
            \item[End Time:] 03:29
            \item[Title:] David's Significance
            \item[Description:] The scene focuses on Michelangelo's David, highlighting its special significance for the people of Florence.
        \end{description}

    \item Scene 9: David in Context
        \begin{description}
            \item[Start Time:] 03:29
            \item[End Time:] 03:42
            \item[Title:] David in Context
            \item[Description:] The scene provides a broader context of Florence, showing the cityscape and comparing different David statues.
        \end{description}

    \item Scene 10: Donatello's David
        \begin{description}
            \item[Start Time:] 03:42
            \item[End Time:] 03:57
            \item[Title:] Donatello's David
            \item[Description:] The scene highlights Donatello's David, emphasizing its historical and artistic significance.
        \end{description}

    \item Scene 11: Michelangelo's David
        \begin{description}
            \item[Start Time:] 03:57
            \item[End Time:] 05:49
            \item[Title:] Michelangelo's David
            \item[Description:] The scene returns to Michelangelo's David, focusing on its artistic details and historical context.
        \end{description}

    \item Scene 12: David's Relocation
        \begin{description}
            \item[Start Time:] 05:49
            \item[End Time:] 06:28
            \item[Title:] David's Relocation
            \item[Description:] The scene discusses the relocation of David to its current position outside the Palazzo della Signoria.
        \end{description}

    \item Scene 13: Historical Figures
        \begin{description}
            \item[Start Time:] 06:28
            \item[End Time:] 06:42
            \item[Title:] Historical Figures
            \item[Description:] The scene highlights historical figures like Piero de' Medici and Savonarola, emphasizing their impact on Florence.
        \end{description}
        
    \item Scene 14: Michelangelo's David
        \begin{description}
            \item[Start Time:] 06:42
            \item[End Time:] 06:59
            \item[Title:] Michelangelo's David
            \item[Description:] The scene focuses on Michelangelo's David as a symbol of the newly reconstituted Republic of Florence.
        \end{description}

\end{enumerate}

\\
\bottomrule
\end{tabular}

\begin{tabular}{l p{12cm}}
    \toprule
    \multicolumn{2}{c} {\textbf{\name~Example - Long Video Understanding (Scene Splitting)}} \\
    \midrule
\textbf{\name} &
(continued) 

\begin{enumerate}[start=15]

    \item Scene 15: Accademia Gallery
        \begin{description}
            \item[Start Time:] 06:59
            \item[End Time:] 07:03
            \item[Title:] Accademia Gallery
            \item[Description:] The scene returns to the Accademia Gallery, showcasing the statue of David and its historical significance.
        \end{description}

    \item Scene 16: Cathedral of Florence
        \begin{description}
            \item[Start Time:] 07:03
            \item[End Time:] 07:10
            \item[Title:] Cathedral of Florence
            \item[Description:] The scene shifts to the Cathedral of Florence, highlighting its architectural significance.
        \end{description}

    \item Scene 17: David's Significance
        \begin{description}
            \item[Start Time:] 07:10
            \item[End Time:] 07:20
            \item[Title:] David's Significance
            \item[Description:] The scene emphasizes the significance of Michelangelo's David in the context of Renaissance Florence.
        \end{description}

    \item Scene 18: Conclusion
        \begin{description}
            \item[Start Time:] 07:20
            \item[End Time:] 07:24
            \item[Title:] Conclusion
            \item[Description:] The video concludes with the \textit{smarthistory} logo, summarizing the key points discussed.
        \end{description}

    \item Scene 19: Credits
        \begin{description}
            \item[Start Time:] 07:24
            \item[End Time:] 07:27
            \item[Title:] Credits
            \item[Description:] The scene provides credits for the video, including contributors and sources.
        \end{description}
\end{enumerate}

The video has a total of 19 scenes, each with specific start and end times, titles, and detailed descriptions.

\\
\bottomrule
\end{tabular}
\footnote{Video source: \url{https://youtube.com/watch?v=QdlP8ai8trw}.}

\pagebreak

\begin{tabular}{l p{12cm}}
    \toprule
    \multicolumn{2}{c}{\textbf{\name~Example - Code Understanding from Videos}}\\ 
    \midrule
    \textbf{User} & On the page with text `Predict 10 Labels` on it, a Python class is defined. Please note down the code that defines this class. After that, explain each part of the code.\\
& \begin{center}
\includegraphics[width=0.8\textwidth]{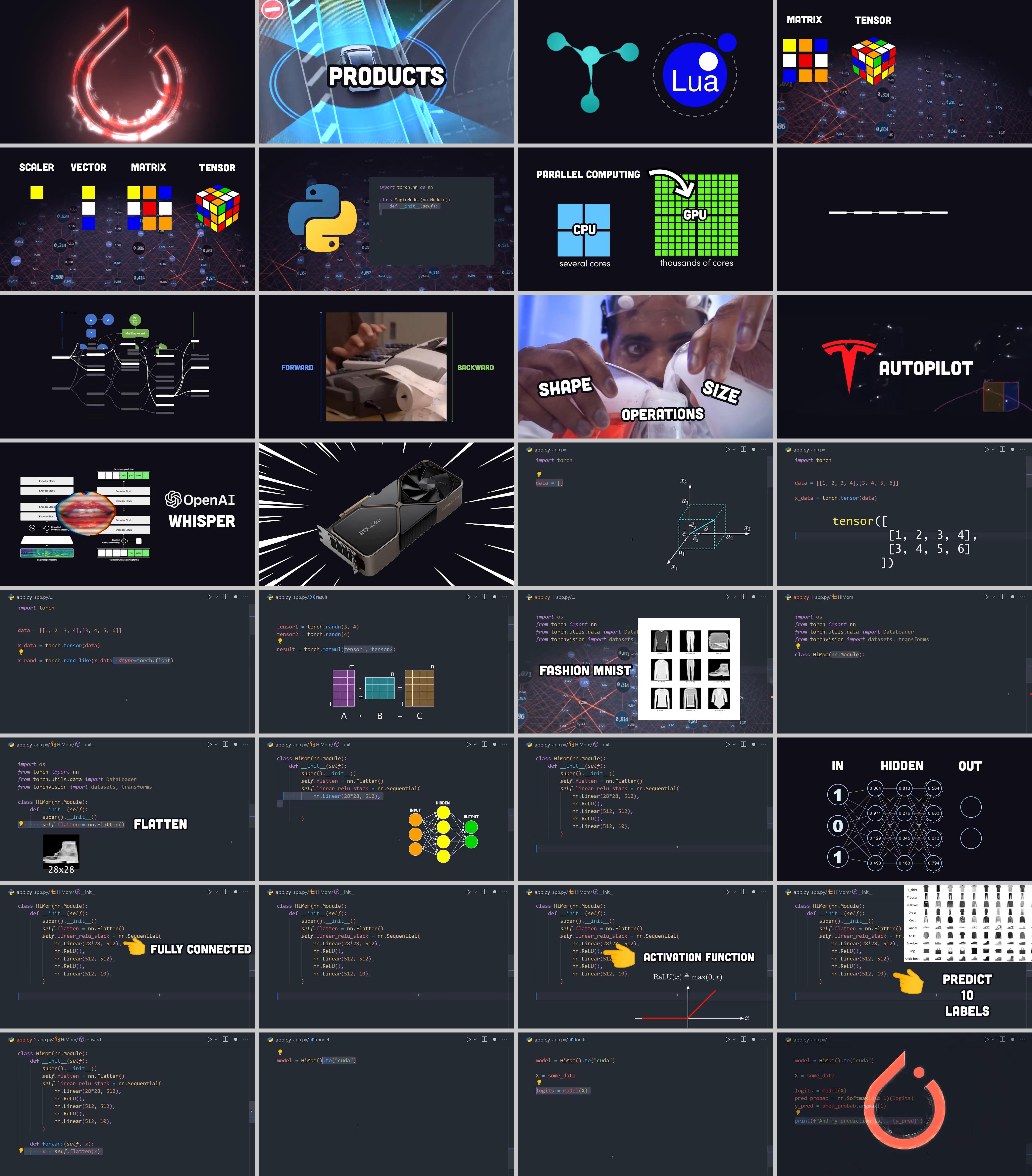}    
\end{center}
 \\
\midrule
\textbf{\name} & The Python class defined on the page with text Predict 10 Labels is as follows:

\begin{lstlisting}
class HiMom(nn.Module):
    def __init__(self):
        super().__init__()
        self.flatten = nn.Flatten()
        self.linear_relu_stack = nn.Sequential(
            nn.Linear(28*28, 512),
            nn.ReLU(),
            nn.Linear(512, 512),
            nn.ReLU(),
            nn.Linear(512, 10),
        )
    def forward(self, x):
        x = self.flatten(x)
        logits = self.linear_relu_stack(x)
        return logits
\end{lstlisting}

\\
\bottomrule
\end{tabular}

\begin{tabular}{l p{12cm}}
    \toprule
    \multicolumn{2}{c} {\textbf{\name~Example - Code Understanding from Videos}} \\
    \midrule
\textbf{\name} &
(continued) Explanation of the code:

\begin{enumerate}
    \item \textbf{Class Definition}: The class \texttt{HiMom} is defined, inheriting from \texttt{nn.Module}.
    
    \item \textbf{Initialization (\_\_init\_\_)}:
    \begin{itemize}
        \item {\small \texttt{super().\_\_init\_\_()}}: Calls the initializer of the parent class.
        \item {\small \texttt{self.flatten = nn.Flatten()}}: Creates a flatten layer to convert input tensors to 1D tensors.
        \item {\small \texttt{self.linear\_relu\_stack = nn.Sequential(...)}}: Defines a sequence of layers:
        \begin{itemize}
            \item {\small \texttt{nn.Linear(28*28, 512)}}: A fully connected layer with 28*28 input features and 512 output features.
            \item {\small \texttt{nn.ReLU()}}: ReLU activation function.
            \item {\small \texttt{nn.Linear(512, 512)}}: Another fully connected layer with 512 input and output features.
            \item {\small \texttt{nn.ReLU()}}: ReLU activation function.
            \item {\small \texttt{nn.Linear(512, 10)}}: Output layer with 10 output features.
        \end{itemize}
    \end{itemize}
    
    \item \textbf{Forward Method}:
    \begin{itemize}
        \item {\small \texttt{x = self.flatten(x)}}: Flattens the input tensor.
        \item {\small \texttt{logits = self.linear\_relu\_stack(x)}}: Passes the flattened tensor through the defined sequence of layers.
        \item {\small \texttt{return logits}}: Returns the output logits.
    \end{itemize}
\end{enumerate}
\\
\bottomrule
\end{tabular}
\footnote{Video source: \url{https://www.youtube.com/watch?v=ORMx45xqWkA}.}

\newpage

\section{Acknowledgement}

The authors appreciate Tianhang Zhu, Chenxu Lv, Xiang He, Shenzhi Wang, Chujie Zheng, Wenhao Huang, Ziyang Luo, Shiming Yang, Fengji Zhang, Yuhao Yang, Linli Yao, Xiao Li, Liying Li, Wen Xie, Peng Liu and Jun Tian for their valuable input and suggestions. 

\bibliographystyle{abbrvnat}
\bibliography{main}

\end{document}